\documentclass{article}
\usepackage{ijcai26}

\usepackage{times}
\usepackage{soul}
\usepackage{url}
\usepackage[hidelinks]{hyperref}
\usepackage[utf8]{inputenc}
\usepackage[small]{caption}
\usepackage{graphicx}
\usepackage{amsmath}
\usepackage{amsthm}
\usepackage{booktabs}
\usepackage{algorithm}
\usepackage{algorithmic}
\usepackage[switch]{lineno}
\usepackage{bm}
\usepackage{amsmath}
\usepackage{amssymb}
\usepackage{multirow}
\usepackage{makecell}

\title{Refining Strokes by Learning Offset Attributes between Strokes\\ for Flexible Sketch Edit at Stroke-Level}

\author{
Sicong Zang\thanks{Corresponding Author.} \and
Tao Sun \and
Cairong Yan
\affiliations
School of Information and Intelligent Science, Donghua University\\
\emails
sczang@dhu.edu.cn\and 2242905@mail.dhu.edu.cn\and cryan@dhu.edu.cn
}

\begin{document}

\maketitle

\begin{abstract}
    Sketch edit at stroke-level aims to transplant source strokes onto a target sketch via stroke expansion or replacement, while preserving semantic consistency and visual fidelity with the target sketch. Recent studies addressed it by relocating source strokes at appropriate canvas positions. However, as source strokes could exhibit significant variations in both size and orientation, we may fail to produce plausible sketch editing results by merely repositioning them without further adjustments. For example, anchoring an oversized source stroke onto the target without proper scaling would fail to produce a semantically coherent outcome. In this paper, we propose SketchMod to refine the source stroke through transformation so as to align it with the target sketch's patterns, further realize flexible sketch edit at stroke-level. As the source stroke refinement is governed by the patterns of the target sketch, we learn three key offset attributes (scale, orientation and position) from the source stroke to another, and align it with the target by: 1) resizing to match spatial proportions by scale, 2) rotating to align with local geometry by orientation, and 3) displacing to meet with semantic layout by position. Besides, a stroke's profiles can be precisely controlled during sketch edit via the exposed captured stroke attributes. Experimental results indicate that SketchMod achieves precise and flexible performances on stroke-level sketch edit.
\end{abstract}

\section{Introduction}

\begin{figure}[!t]
    \centering
    \includegraphics[width=\columnwidth]{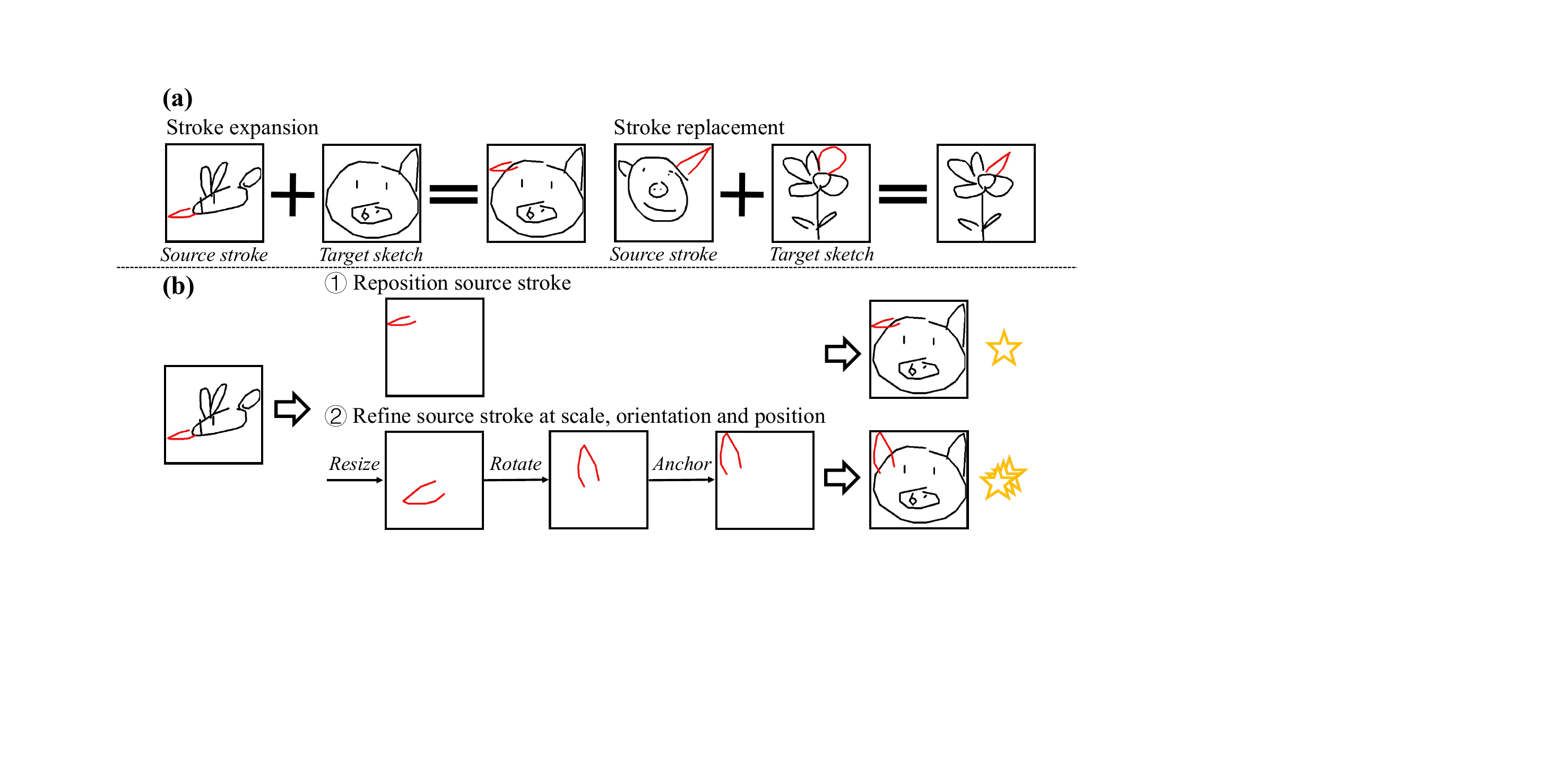}
    \caption{(a) Two applications of stroke-level stroke edit. Stroke expansion: Adding the source stroke in red onto the target sketch. Stroke replacement: Replacing the red stroke in target by the source stroke. (b) Two approaches to refine source stroke for stroke expansion. 1) Predicting a new position to locate, which is utilized by SketchEdit \protect\cite{li2024sketchedit} and Sketch-HARP \protect\cite{zang2025generating}. 2) Refining source stroke at scale, orientation and position, allowing it to align with the target sketch's patterns for flexible sketch edit.}
    \label{fig:sketch_edit}
\end{figure}

Sketch edit at stroke-level targets to 1) transplant source strokes onto the target sketch, namely stroke expansion, or 2) replacing the selected strokes form the target with source strokes, namely stroke replacement. Two examples are shown in Fig.~\ref{fig:sketch_edit}(a). The generated sketch via sketch edit is required to maintain semantic consistency and visual fidelity with the target, while ensuring source strokes align with the target sketch's patterns. It is a challenging task as models are required to decide where and how to anchor source strokes onto the target by comprehending not only the global patterns from the canvas, but also the local features of what the source strokes represent.

Recently, SketchEdit \cite{li2024sketchedit} and Sketch-HARP \cite{zang2025generating} are proposed for addressing stroke-level sketch edit by relocating source strokes at appropriate positions on the canvas. For example, shown as the first approach in Fig.~\ref{fig:sketch_edit}(b), the bee stinger (as a source stroke) is positionally adjusted to be placed over the pig's face (as the target sketch) to disguise itself as the pig's right ear. However, as source strokes could exhibit significant variations in many attributes such as size and orientation, sometimes we may fail to produce plausible sketch editing results by merely repositioning them without further adjustments. For instance, anchoring an oversized bee stinger onto a pig's face without proper scaling would fail to produce a semantically coherent outcome. The attribute misalignment between the source stroke and the target sketch would reduce both the quality of sketch editing and the adaptability of the editing process.

Effective alignment between source strokes and the target sketch's patterns can be achieved through transformative adjustment at three key attributes of strokes: scale, orientation, and position. The second approach in Fig.~\ref{fig:sketch_edit} reports an example to reveal how these attributes could be adjusted cooperatively to improve sketch edit. More specifically, scale determines the size of a stroke. An over-/under-sized source stroke can be rescaled to fit the strokes' sizes in the target sketch. Orientation controls the direction of a stroke. Rotating a stroke changes both its visual appearance and semantic representation, which might assist its alignment towards the target's patterns. By refining a source stroke at scale and orientation, it would be more flexible to be anchored at an appropriate location to obtain satisfied sketch edit results. And as a result in Fig.~\ref{fig:sketch_edit}(b), the new right ear from the refined bee stinger is much more recognizable to compose the pig face.

The refinement of source strokes is governed by the patterns of the target sketch. More specifically, the attributes from the target sketch define three critical alignment criteria for the source strokes: 1) resizing to match spatial proportions by \emph{scale}, 2) rotating to align with local geometry by \emph{orientation}, and 3) displacing to meet with semantic layout by \emph{position}. These alignments are guided by offset attributes, which quantify the relative displacements between source and target strokes across these three attributes. Offset attribute is a kind of directed relative position to provide spatial and structural information from one stroke to another. We equip stroke features with them and employ message passing networks to enable the communication from the target towards the source strokes, enabling adaptive refinement that respects both local (stroke-level) and global (sketch-level) coherence.

To realize these above, we propose SketchMod for flexible stroke-level sketch edit via stroke refinement by learning offset attributes between strokes. By given a source stroke and a target sketch, SketchMod firstly captures all stroke embeddings and predicts their stroke attributes (including position, scale and orientation), which are utilized to learn relative offset attributes between the source stroke and another one from the target. Features from normalized strokes are aggregated by their relative offsets to obtain a refined source stroke, whose stroke attributes are then captured and coordinate with features from the normalized strokes to generate sketch edit results. To summarize, we make the following contributions:
\begin{itemize}
	\item[1.] We propose SketchMod for flexible stroke-level sketch edit via refining source strokes at scale, orientation and position.
	\item[2.] We equip strokes with relative positions among stroke, which are quantified by their corresponding offset attributes, for source stroke refinement via message aggregation.
	\item[3.] Experimental results indicate that SketchMod achieves accurate and flexible performance on stroke-level sketch edit, e.g., stroke replacement and expansion.
\end{itemize}

\begin{figure*}[!t]
    \centering
    \includegraphics[width=1.8\columnwidth]{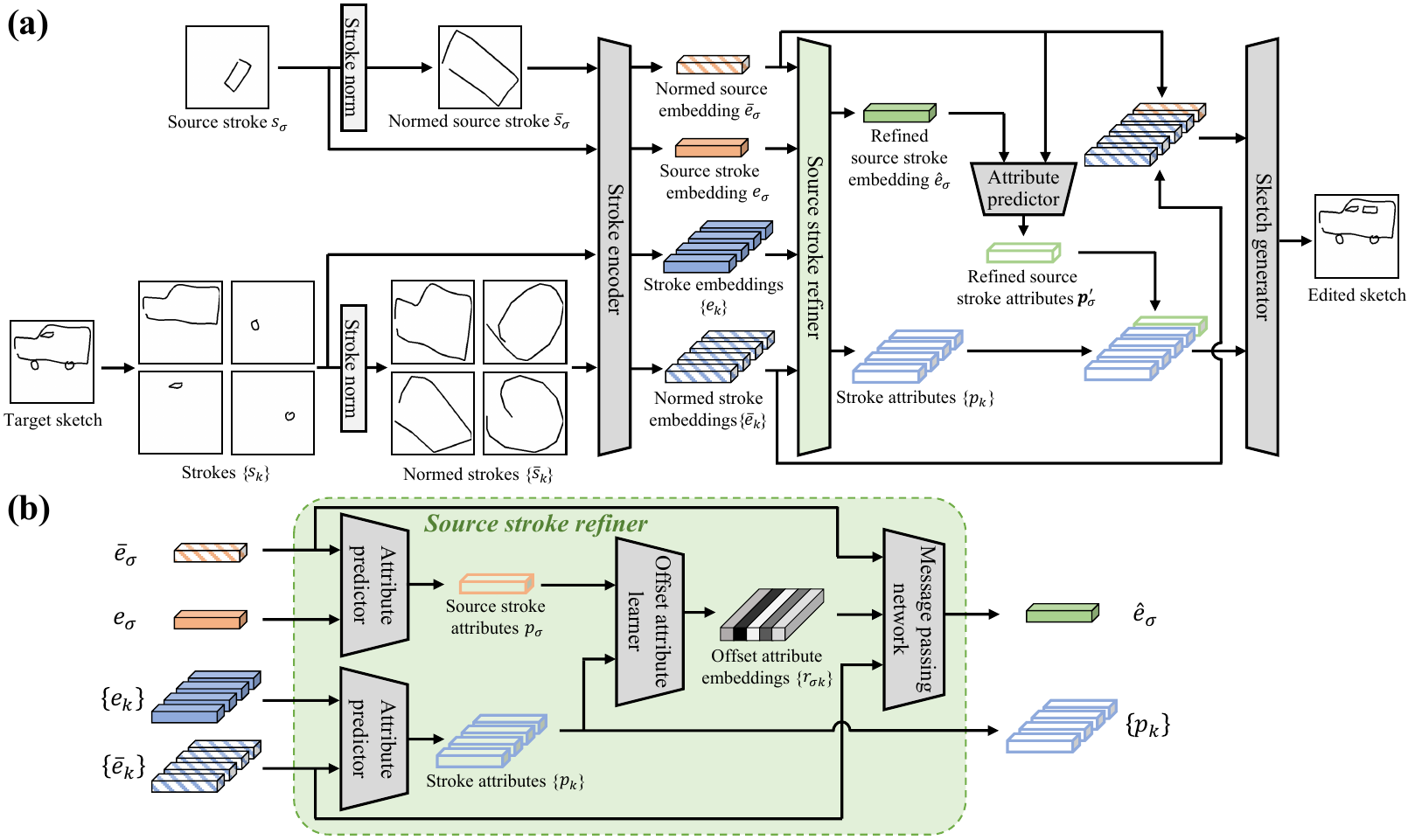}
    \caption{Applying stroke-level sketch edit by SketchMod. (a) The network structure. After capturing embeddings of source stroke and the ones from the target sketch, we learn offset attributes between them to obtain a refined source stroke embedding, whose attributes are predicted and fed into a sketch generator along with all normalized strokes to generate the edited sketch. (b) Architecture of the source stroke refiner. Stroke attributes are predicted from stroke, and are utilized to learn relative offset attribute embeddings between source stroke and another. A message passing network is employed to aggregate messages from all normalized strokes by their relative offsets to finally obtain the refined source stroke embedding.}
    \label{fig:overview}
\end{figure*}

\section{Related Work}

\subsection{Controllable Sketch Synthesis}

Controllable sketch synthesis requires models to generate sketches with specific categorical and stylistic patterns as expected. The stylistic patterns are some non-categorical features, e.g., the orientation of giraffes, which are critical to describe detailed patterns for sketch representation. The performance reveals the ability about whether a model could generate sketches accurately and robustly. 

Recent applications on controllable sketch synthesis mainly focuses on manipulating a sketch's global patterns through latent codes, rasterized (sketch) images, prompts, etc. Sketch healing \cite{su2020sketchhealer,qi2022generative} requires a model to restore corrupted sketches by predicting their missing details under masks. Sketch reorganization \cite{wang2024self} rectifies the misplacement of strokes by repositioning them on the canvas. Sketch analogy \cite{ha2017neural,zang2024self} combines patterns collected from multiple sketches to generate hybrid output. Text-guided sketch refinement \cite{tian2025sketchrefiner} enhances rough free-hand sketches into detailed renderings by using text prompts. These models fail to precisely modify a specific stroke's patterns without altering the remaining strokes' patterns.

This paper advances a stroke-specified controllable sketch synthesis task, namely stroke-level sketch edit, by refining the source stroke to align it with the target sketch's patterns.

\subsection{Stroke-Level Sketch Edit}

Sketch edit at stroke-level transplants some given source strokes onto a target sketch. The source strokes can be directly transplanted onto the canvas, i.e., stroke expansion, or can be utilized to replace another stroke selected from the target, namely stroke replacement. Both applications aim to generate edited sketches by comprehending both the global patterns from target sketch and the local features contained by the source stroke.

SketchXAINet \cite{qu2023sketchxai} learns to relocate a sequence of source strokes, which are randomly positioned on a blank canvas, to yield a categorically recognizable sketch. Note that all strokes are fixed at their curve shapes. SketchEdit \cite{li2024sketchedit} addresses stroke-level sketch edit by repositioning source strokes onto the target sketch. Compared with SketchXAINet, SketchEdit fixes source strokes at their features instead of their raw data (drawing actions). It requires SketchEdit to realize what the target sketch and the source stroke are in geometric shape and semantics before drawing it on target sketch. Sketch-HARP \cite{zang2025generating} is able to adjust a stroke's position at any middle stage of sketch generation. Contextual information about how strokes are drawn stroke-by-stroke are learned via a hierarchical auto-regressive generating process to realize sketch drawing manipulation at stroke-level.

We refine the source stroke at scale, orientation and position to make it align with the target sketch's patterns, contributing to flexible sketch edit at stroke-level.

\section{Methodology}

Fig.~\ref{fig:overview} presents an overview of SketchMod. By given a source stroke and a target sketch, SketchMod firstly captures all stroke embeddings and predicts their corresponding attributes (including position, scale and orientation), which are further utilized to learn relative offset attributes between source stroke and another. Features from the normalized strokes are aggregated by their relative offsets to obtain a refined source stroke, which aligns with the target sketch's patterns. A sketch generator collects predicted attributes of the refined strokes along with their corresponding normalized strokes to generate the edited sketch.

\subsection{Offset Attributes between Strokes} \label{sec:stroke_norm}

\begin{figure}[!t]
    \centering
    \includegraphics[width=0.4\columnwidth]{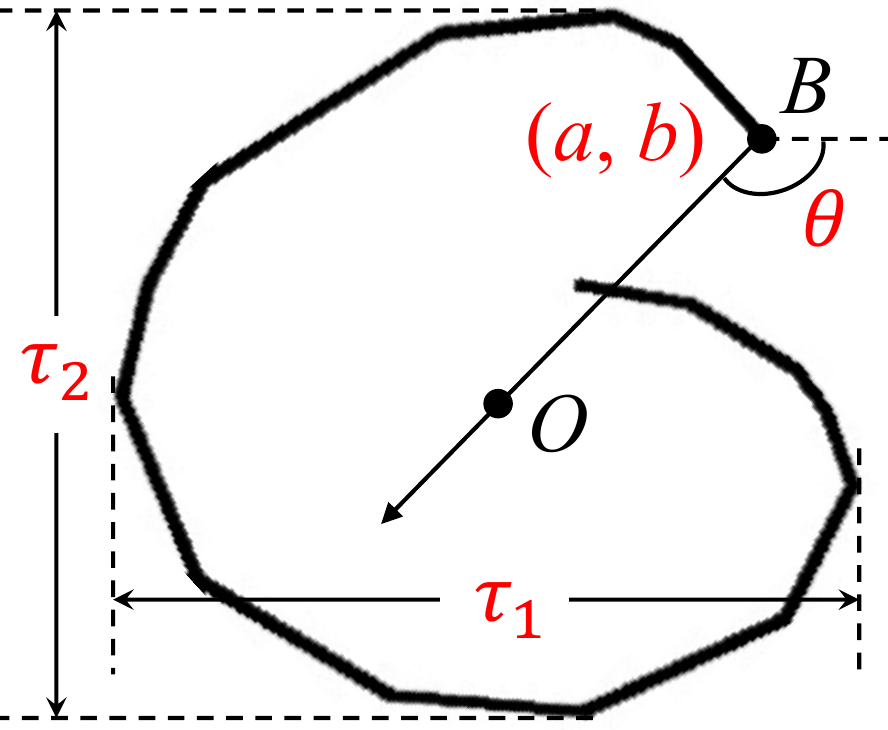}
    \caption{The definitions of three stroke attributes: position $(a, b)$, orientation $\theta$ and scale $\bm\tau=[\tau_1, \tau_2]$.}
    \label{fig:attributes}
\end{figure}

\textbf{Editable stroke attributes}. When editing sketches at stroke-level, three key stroke attributes could be adjusted to align source stroke with the target sketch in both visual and semantic patterns. Fig.~\ref{fig:attributes} reveals their definitions.

\emph{Position}. This attribute determines where a stroke is anchored on the canvas. We use the 2D coordinate of a stroke's starting point, e.g., $(a, b)$ for the starting point \emph{B} in Fig.~\ref{fig:attributes}, to represent its position. Locating a stroke at an appropriate position contributes to recognizable sketches. And recent methods, e.g., SketchEdit \cite{li2024sketchedit} and Sketch-HARP \cite{zang2025generating}, predict starting position of the source stroke to benefit sketch edit.

\emph{Orientation}. This attribute indicates the direction in which a stroke is facing. As in Fig.~\ref{fig:attributes}, for a stroke with a starting point \emph{B} and a geometric center \emph{O}, its orientation is defined by the direction of $\overrightarrow{BO}$, and we quantify it by the angle $\theta$. The adjustment of a stroke's orientation can be represented as applying a rotational transformation about the reference point \emph{B}.

\emph{Scale}. This attribute controls the size of a stroke. Scale is a two-dimensional vector with two elements to determine the sizes along x- or y-axis, respectively, e.g., $\bm \tau=[\tau_1, \tau_2]$ in Fig.~\ref{fig:attributes}. Rescaling a source stroke before drawing it on the target sketch could preserve visual coherence by aligning its size with the target sketch's patterns.

In practice, for a stroke $\bm s=[\bm x, \bm y, \bm l]$\footnote{In QuickDraw \cite{ha2017neural}, each stroke consists of a sequence of drawing actions, i.e., $\bm s=[\bm x, \bm y, \bm l]$, where $(\bm x, \bm y)$ denotes the 2D coordinates on the pen's movement and $\bm l$ is a one-hot vector indicating the pen states (pen lifting, pen down or end-of-drawing).}, we capture its attributes by normalizing the stroke curve via Eq.~(\ref{eq:normalization}).
\begin{align}
    \begin{bmatrix}
            \bm x^{\prime}\\
            \bm y^{\prime}
        \end{bmatrix}&=\begin{bmatrix}
            \cos\theta & -\sin\theta \\
            \sin\theta & \cos\theta
        \end{bmatrix}
    \cdot\begin{bmatrix}
            \bm x-a\\
            \bm y-b
        \end{bmatrix},\nonumber\\
    [\bar{\bm x}, \bar{\bm y}]&=\frac{
        [\bm x^{\prime}, \bm y^{\prime}]
    -[\min(\bm x^{\prime}), \min(\bm y^{\prime})]}
    {[\max(\bm x^{\prime}), \max(\bm y^{\prime})]
    -[\min(\bm x^{\prime}), \min(\bm y^{\prime})]},
    \label{eq:normalization}
\end{align}
where $\min(\cdot)$ and $\max(\cdot)$ denote the minimum and the maximum operation, respectively. More specifically, the starting point $(a, b)$ of a stroke is captured to represent its position firstly. And then we rotate the stroke by an angle of rotation $\theta$ to align it with a pre-defined reference direction. Finally, we resize the stroke by a min-max normalization to obtain its corresponding normalized stroke $\bar{\bm s}=[\bar{\bm x}, \bar{\bm y}, \bm l]$ and the scale vector $\bm\tau=[\max(\bm x-a)-\min(\bm x-a), \max(\bm y-b)-\min(\bm y-b)]$. Note that we use the log-scale $\log(\bm\tau)$ to represent a stroke's size, which will be explained in the following paragraph. In summary, three editable attributes are represented by a five-dimensional vector $\bm p=[a, b, \theta, \log(\bm\tau)]$.

\textbf{Offset attributes between strokes}. When realizing flexible stroke-level sketch edit, it is important to consider the offset attributes between the source stroke and ones in target sketch, as it is beneficial to align source stroke with the target's patterns at position, orientation and scale, ensuring that the edited sketches are recognizable in appearance and reasonable in semantics. Since we quantify three attributes of a stroke, the relationships between strokes, more specifically, their offset attributes, can be easily modeled. For two strokes $\bm s_i$ and $\bm s_j$, the offset position from $\bm s_j$ to $\bm s_i$ can be computed as $(a_i-a_j, b_i-b_j)$, revealing the Euclidean displacement from $\bm s_j$ to $\bm s_i$. Similarly, the offset orientation from $\bm s_j$ to $\bm s_i$ is $\theta_i-\theta_j$, describing the angular displacement from $\bm s_j$'s direction to $\bm s_i$'s direction. Since the stroke size is represented in logarithmic form as $\log(\bm\tau)$, the relative scale offset from $\bm s_j$ to $\bm s_i$, which is typically quantified by their size ratio, can be expressed as $\log(\bm\tau_i)-\log(\bm\tau_j)=\log(\bm\tau_i/\bm\tau_j)$. In summary, we obtain the offset attributes from $\bm s_j$ to $\bm s_i$ by $\bm p_i-\bm p_j$.

\subsection{Refining Source Strokes by Learning Offset Attributes between Strokes}

In order to improve sketch edit by aligning the source stroke $\bm s_\sigma$ with the patterns of the target sketch with $K$ strokes $\{\bm s_k\}_{k=1}^K$, $\bm s_\sigma$ is encouraged to be refined in an appropriate manner at its scale, orientation and position by an introduced source stroke refiner, shown in Fig.~\ref{fig:overview}(b).

In practice, we firstly normalize $\bm s_\sigma$ and $\{\bm s_k\}$ via Eq.~(\ref{eq:normalization}) to obtain their corresponding normalized strokes $\bar{\bm s}_\sigma$ and $\{\bar{\bm s}_k\}$. A multi-layer perception (MLP)-based stroke encoder $f$ is introduced to capture their stroke embeddings $\bm e\in\mathbb{R}^{128\times 1}$,
\begin{align}
    \bm e_\sigma=f(\bm s_\sigma),\>\bar{\bm e}_\sigma=f(\bar{\bm s}_\sigma),\>\bm e_k=f(\bm s_k),\>\bar{\bm e}_k=f(\bar{\bm s}_k).
    \label{eq:stroke_enc}
\end{align}

The normalized stroke embeddings $\bar{\bm e}_\sigma$ and $\{\bar{\bm e}_k\}$ contain features to represent the generalized shape of strokes, e.g., whether it is a line, a square or a circle, to categorize their intrinsic geometric properties. And $\bm e_\sigma$ and $\{\bm e_k\}$ from the original strokes store more information about the detailed attributes of the curve, e.g., whether it is a circle or an ellipse. We introduce an attribute predictor $\mathcal{F}$ to predict the editable attributes by feeding it with embedding pairs $\bm e$ and $\bar{\bm e}$.
\begin{align}
    \bm p_\sigma=\mathcal{F}(\text{CAT}(\bm e_\sigma; \bar{\bm e}_\sigma)),\>\bm p_k=\mathcal{F}(\text{CAT}(\bm e_k; \bar{\bm e}_k)),
    \label{eq:abs_enc}
\end{align}
where $\text{CAT}(\cdot)$ denotes the concatenation operation. $\mathcal{F}$ consists of three linear layers, with the first two followed by the layer normalization and the GeLU activation. The learned $\bm p$ is a five-dimensional vector with the predicted stroke attributes $[a, b, \theta, \log(\bm \tau)]$. According to Sect.~\ref{sec:stroke_norm}, we can easily represent the offset attributes from stroke $\bm s_k$ to $\bm s_\sigma$ by computing $\bm p_\sigma-\bm p_k$. We further employ an MLP, noted as $h$, to learn the offset attribute embedding $\bm r_{\sigma k}\in\mathbb{R}^{128\times 1}$.
\begin{align}
    \bm r_{\sigma k}=h(\bm p_\sigma-\bm p_k).
    \label{eq:rel_enc}
\end{align}

With the assistance of $\bm r_{\sigma k}$, we could model the directed relationships from the target sketch to the source stroke. By further integrating messages from $\{\bm s_k\}$, we refine $\bm s_\sigma$ to yield an adapted source stroke that better aligns with the target sketch's patterns. Practically, we introduce a three-layer stacked message passing network $\psi$ to obtain the refined source stroke $\hat{\bm e}_\sigma$.
\begin{align}
    \hat{\bm e}_\sigma=\psi(\bar{\bm E}, \bm R),
    \label{eq:predictor}
\end{align}
where $\bar{\bm E}=\text{CAT}(\bar{\bm e}_\sigma; \{\bar{\bm e}_k\})$, and $\bm R\in\mathbb{R}^{128\times (K+1)\times (K+1)}$ collects all the offset attribute embeddings with $\bm R_{ij}=\bm r_{ij}$.

The detailed architecture of $\psi$ is shown in Eq.~(\ref{eq:message})-Eq.~(\ref{eq:ffn}).
\begin{align}
    \bm A^{(t)}_{ij}=&\bar{\bm E}_i^{(t)\top}\bm W_q^{(t)\top}\bm W_{k,E}^{(t)} \bar{\bm E}^{(t)}_j+\bar{\bm E}_i^{(t)\top}\bm W_q^{(t)\top}\bm W_{k, R}^{(t)}\bm R_{ij}\nonumber\\
    &+\bm u^{\top}\bm W_{k,E}^{(t)}\bar{\bm E}^{(t)}_j+\bm v^{\top}\bm W_{k,R}^{(t)}\bm R_{ij},\label{eq:message}\\
    \bar{\bm H}^{(t+1)}_i=&\sum\nolimits_j\mathop{\text{Softmax}}\limits_j(\bm A^{(t)}_{ij})\cdot(\bm W_v^{(t)}\bar{\bm E}^{(t)}_j+\bm R_{ij}),\label{eq:aggregate}\\
    \bm\Phi^{(t+1)}=&\text{LayerNorm}(\text{Linear}(\bar{\bm H}^{(t+1)})+\bar{\bm E}^{(t)}),\\
    \bar{\bm E}^{(t+1)}=&\text{FFN}(\bm\Phi^{(t+1)}),\label{eq:ffn}
\end{align}
where the superscript $(t)$ denotes the $t$-th ($t=1,2,3$) message passing layer. $\bm W_q$, $\bm W_{k,E}$, $\bm W_{k,R}$, $\bm W_v\in\mathbb{R}^{128\times 128}$ and $\bm u$, $\bm v\in\mathbb{R}^{128\times 1}$ are learnable parameters. $\text{FFN}(\cdot)$ indicates a feed forward network implemented by an MLP.

Specifically, the captured offset attributes are regarded as information about a kind of relative positions. Inspired by \cite{dai2019transformer}, we decompose the attention score $\bm A_{ij}$ by four terms in Eq.~(\ref{eq:message}). The four terms represent: 1) an addressing based on normalized stroke contents, 2) a stroke content dependent attribute offset, 3) a global content bias and 4) a global attribute offset, respectively. Following \cite{wu2021rethinking}, we also collect the relative positions $\bm R_{ij}$ during the message aggregation in Eq.~(\ref{eq:aggregate}) to cooperate with information from the normalized strokes $\bar{\bm E}_j$. Note that information from the original source stroke $\bm s_\sigma$ does not participate in its refinement, as its original stroke attributes may corrupt its refinement. Instead, only the offset attributes are utilized to achieve its alignment.

By collecting the refined source stroke embedding $\hat{\bm e}_\sigma$ from $\psi$, we feed it into an attribute predictor $\mathcal{F}$, which shares the same parameters from the one in Eq.~(\ref{eq:abs_enc}), along with its normalized stroke $\bar{\bm e}_\sigma$ to yield the new stroke attributes $\bm p^{\prime}_\sigma$ from the refined stroke.
\begin{align}
    \bm p^{\prime}_\sigma=\mathcal{F}(\text{CAT}(\hat{\bm e}_\sigma;\bar{\bm e}_\sigma)).
\end{align}

\subsection{Generating the Edited Sketch}

For a source stroke, we have captured its normalized stroke embedding $\bar{\bm e}_\sigma$ and have learned its new attributes $\bm p^{\prime}_\sigma$ from the corresponding refined stroke. We equip each normalized stroke embedding with its stroke attributes via addition, i.e., $\bar{\bm e}+\bm p$, and concatenate it with the ones from the target sketch, and send them into a four-layer stacked self-attention network $\xi$ to enhance their interactions to further yield comprehensive stroke embeddings $\tilde{\bm e}_\sigma, \{\tilde{\bm e}_k\}$.
\begin{align}
    \tilde{\bm e}_\sigma, \{\tilde{\bm e}_k\}=\xi(\text{CAT}(\bar{\bm e}_\sigma+\bm p^{\prime}_\sigma; \{\bar{\bm e}_k+\bm p_k\})).
    \label{eq:mixing_stroke}
\end{align}
These final embeddings $\tilde{\bm e}_\sigma, \{\tilde{\bm e}_k\}$ are respectively sent into two generators, adopted from SketchEdit \cite{li2024sketchedit}, to produce sequence- and image-formed sketches.
\begin{align}
    \widehat{seq}=\mathcal{G}_{\text{seq}}(\tilde{\bm e}_\sigma, \{\tilde{\bm e}_k\}),\>\hat{\bm I}=\mathcal{G}_{\text{img}}(\tilde{\bm e}_\sigma, \{\tilde{\bm e}_k\}).
    \label{eq:dec}
\end{align}

\subsection{Training a SketchMod}

Before training a SketchMod, we separate each sketch from the training set into a pair of a source stroke and a target sketch, and regard the original sketch as a ground truth sketch edit result. In practice, for an input sketch with $K+1$ strokes, we randomly select a stroke as the source stroke $\bm s^{\text{GT}}_\sigma$ and the rest ones regarded as the target sketch composed by $K$ strokes $\{\bm s_k\}_{k=1}^K$. After capturing the ground truth attributes $\bm p^{\text{GT}}_\sigma$ from $\bm s^{\text{GT}}_\sigma$, we apply random noises over the attributes of $\bm s^{\text{GT}}_\sigma$ to obtain a corrupted stroke $\bm s_\sigma$, which is utilized as the given source stroke for sketch edit. SketchMod has two training stages (more details in Appendix A):

\textbf{Stage I: For sketch generation}. In stage I, we train the stroke encoder $f$, the attribute predictor $\mathcal{F}$ and the sketch generators (parameterized by $\xi, \mathcal{G}_\text{seq}$ and $\mathcal{G}_\text{img}$) via sketch reconstruction. Without corrupting the source stroke, all strokes from a complete sketch are fed into the stroke encoder to extract their embeddings, which bypass the source stroke refiner but are utilized to predict their stroke attributes. All the predicted attributes with their corresponding normalized stroke embeddings are proceeded to the sketch generator for sketch reconstruction. The objective in this stage is minimizing,
\begin{equation}
    \resizebox{\hsize}{!}{$
    \min\limits_{f, \mathcal{F}, \xi, \mathcal{G}_{\text{seq}}, \mathcal{G}_{\text{img}}}\mathcal{L}_1=\mathcal{L}_{\text{seq}}+0.2\cdot\|\bm I-\hat{\bm I}\|_2^2+\sum\nolimits_k\|\bm p^{\text{GT}}_k-\bm p_k\|_2^2,
    $}
    \label{eq:stage1_loss}
\end{equation}
where $\|\cdot\|_2^2$ denotes L2 norm. $\mathcal{L}_{\text{seq}}$ represents the sketch sequence reconstruction, and we compute this term following \cite{ha2017neural}. We also adopt the sketch image reconstruction term as \cite{li2024sketchedit,zang2025generating}, assisting SketchMod's training with visual patterns. The third term regularizes all the predicted attributes $\{\bm p_k\}_{k=1}^{K+1}$ to move towards their targets.

\textbf{Stage II: For source stroke refinement}. In stage II, leveraging the pretrained parameters, we keep the stroke encoder, attribute predictor and sketch generator frozen and refine the input corrupted source stroke $\bm s_\sigma$ to recover its corresponding ground truth $\bm s^{\text{GT}}_\sigma$ by learning the refined stroke embedding $\hat{\bm e}_\sigma$ and its corresponding attributes $\bm p^{\prime}_\sigma$. The objective is minimizing,
\begin{align}
    \min\limits_{h, \psi}\mathcal{L}_2=\|\text{sg}(\bm e^{\text{GT}}_\sigma)-\hat{\bm e}_\sigma\|_2^2+\|\bm p^{\text{GT}}_\sigma-\bm p^{\prime}_\sigma\|_2^2.
    \label{eq:stage2_loss}
\end{align}
These two terms push the refined source stroke embedding $\hat{\bm e}_\sigma$ and its corresponding attributes $\bm p_\sigma^{\prime}$ towards their targets ($\bm e_\sigma^{\text{GT}}$ and $\bm p_\sigma^{\text{GT}}$, respectively) captured from the original stroke $\bm s_\sigma^{\text{GT}}$.

\section{Experiments and Analysis}

\subsection{Preparations}

\textbf{Datasets}. Two datasets from QuickDraw \cite{ha2017neural} are selected following \cite{li2024sketchedit,zang2025generating} to verify the sketch edit performance. Dataset 1 (DS1) contains 17 categories (airplane, angel, alarm clock, apple, butterfly, belt, bus, cake, cat, clock, eye, fish, pig, sheep, spider, umbrella and the Great Wall of China). Sketches in DS1 have recognizable categorical patterns. DS2 covers 5 categories (bee, bus, flower, giraffe and pig), and these sketches have vivid non-categorical patterns, e.g., a pig might have an entire body or not. Each category contains $70K$ training, $2.5K$ valid and $2.5K$ test sketches ($1K=1000$).

\textbf{Baselines}. Eight sketch generative models (RPCL-pix2seq \cite{zang2021controllable}, RPCL-pix2seqH \cite{zang2024self}, SketchHealer \cite{su2020sketchhealer}, Lmser-pix2seq \cite{li2024lmser}, SP-gra2seq \cite{zang2023linking}, DC-gra2seq \cite{zang2025equipping} and SketchEdit \cite{li2024sketchedit} and Sketch-HARP \cite{zang2025generating}) are selected as baselines. SketchEdit and Sketch-HARP are proposed for stroke-level sketch edit, and the rests are for controllable sketch synthesis. Besides, we update SketchEdit by allowing its diffusion-based module to predict the source stroke's scale, orientation and position, and we name this variant as SketchEdit$^+$. 

\begin{table}[!t]
    \centering
    \small
    \begin{tabular}{@{}clcrcc@{}}
        \toprule
        Dataset & Model & $Rec$ & FID & LPIPS & CLIP-S\\
        \midrule
        \multirow{10}*{DS1} & RPCL-pix2seq & 53.03 & 21.89 & 0.39  & 89.52 \\
        ~ & RPCL-pix2seqH & 60.25 & 21.62  & 0.39  & 90.72 \\
        ~ & SketchHealer & 56.04 & 21.13  & 0.42  & 90.33 \\
        ~ & Lmser-pix2seq & 65.37 & 13.89  & 0.37  & 91.78 \\
        ~ & SP-gra2seq & 59.90 & 16.12  & 0.42  & 90.86 \\
        ~ & DC-gra2seq & 63.63 & 9.61  & 0.39  & 92.00 \\
        ~ & SketchEdit & 64.92 & 5.18  & 0.22  & 94.53 \\
        ~ & SketchEdit$^+$ & 63.35 & 6.57  & 0.22  & 94.24 \\
        ~ & Sketch-HARP & 49.25 &10.16  & 0.39  & 91.94 \\
        ~ & SketchMod & \pmb{71.51} & \pmb{3.44} & \pmb{0.19} & \pmb{95.52} \\
        \midrule
        \multirow{10}*{DS2} & RPCL-pix2seq & 44.74 & 40.88 & 0.46 & 86.67\\
        ~ & RPCL-pix2seqH & 49.49 & 32.59 & 0.44 & 88.19 \\
        ~ & SketchHealer & 81.32 & 16.17  & 0.43  & 90.63 \\
        ~ & Lmser-pix2seq & 76.63 & 23.84  & 0.41  & 90.58 \\
        ~ & SP-gra2seq & 82.25 & 11.88  & 0.41  & 91.24 \\
        ~ & DC-gra2seq & 81.34 & 9.48  & 0.40  & 91.82 \\
        ~ & SketchEdit & 78.59 & 7.23  & 0.25  & 93.20 \\
        ~ & SketchEdit$^+$ & 75.14 & 9.58  & 0.25  & 92.88 \\
        ~ & Sketch-HARP & 63.72 & 16.18  & 0.41  & 90.54 \\
        ~ & SketchMod & \pmb{89.25} & \pmb{3.86}  & \pmb{0.17}  & \pmb{95.43} \\
        \bottomrule
    \end{tabular}
    \caption{Comparisons on stroke-level sketch edit.}
    \label{tab:edit}
\end{table}

\textbf{Metrics}. Following \cite{li2024lmser,zang2025generating}, we use $Rec$ \cite{zang2021controllable}, the Fr\'{e}chet Inception Distance (FID) score \cite{heusel2017gans}, the learned perceptual image patch similarity (LPIPS) computed by ControlNet \cite{zhang2023adding} and the contrastive language-image pre-training score (CLIP-S) \cite{hessel2021clipscore} to evaluate sketch edit performances. $Ret$ \cite{zang2021controllable}, which is utilized in \cite{zang2025generating}, is omitted here, as SketchMod does not represent a sketch by a single latent code, making it difficult to evaluate SketchMod and SketchEdit by $Ret$.

\begin{figure*}[!t]
    \centering
    \includegraphics[width=1.8\columnwidth]{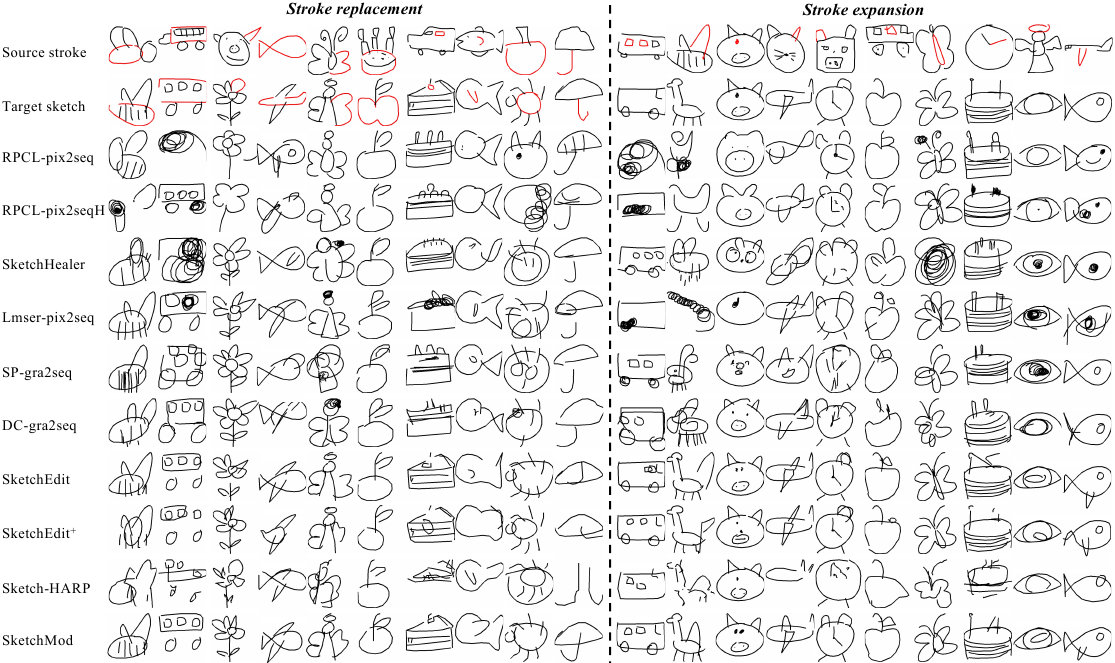}
    \caption{Qualitative comparisons on stroke-level sketch edit. A source stroke highlighted in red is required to be transplanted onto the target sketch via stroke replacement or expansion.}
    \label{fig:edit_results}
\end{figure*}

\subsection{Sketch Edit at Stroke-Level}

We evaluate whether our SketchMod outperforms any other baselines on stroke-level sketch edit. In order to quantify the performance, for each sketch in test set, we randomly select a stroke as the source stroke and corrupt it at scale, orientation and position by adding random noises. The rest strokes are regarded as the target sketch. Each model is required to transplant the source stroke onto the target sketch, and the original complete sketch is as the ground truth result (details in Appendix A). Note that a sketch is always corrupted by the same noises on the same stroke to make the comparison fair. Quantitative results are reported in Table~\ref{tab:edit}.

By adding noises at three key attributes on a stroke, most of the corrupted sketches are difficult to be recognized in their appearance. Accordingly, the models which extracted features from sketch images, e.g., RPCL-pix2seq and RPCL-pix2seqH, fail to learn accurate sketch representations, resulting in poor performances on sketch edit. SketchEdit and Sketch-HARP cannot refine the corrupted source stroke to align with the target sketch's patterns. Despite incorporating a diffusion model for attribute prediction, SketchEdit$^+$ still underperforms our SketchMod. This performance gap demonstrates the critical advantage of our source stroke refining approach via learning offset attributes.

To convincing the quantitative results, we further administer a questionnaire to investigate how human subjects rate sketch edit results. The questionnaire includes 10 stroke replacements and 10 stroke expansions questions (details in Appendix B). Each human subject was required to rank the sketches edited by ten models. Following \cite{zang2025generating}, in each question, only the best three models would be valued at 3, 2 and 1, respectively, leaving the rest models with 0 values. Totally 42 volunteers participated in the evaluation,
and the scoring results in Table~\ref{tab:human_scoring} report that SketchMod is the winner. Further, we raise t-tests in which the p-values against our SketchMod with SketchEdit (the rival in Table~\ref{tab:human_scoring}) are $2.45\times10^{-11}$ and $0.01$ for stroke expansion and replacement, respectively, which reveals that the improvement yielded by SketchMod with SketchEdit is significant.

\begin{table}[!t]
    \centering
    \small
    \begin{tabular}{lcc}
        \toprule
        Model & Expansion & Replacement\\
        \midrule
        RPCL-pix2seq & $0.42\pm0.82$ & $0.34\pm0.72$\\
        RPCL-pix2seqH & $0.37\pm0.76$ & $0.35\pm0.77$\\
        SketchHealer & $0.37\pm0.78$ & $0.22\pm0.64$\\
        Lmser-pix2seq & $0.18\pm0.55$ & $0.77\pm0.91$\\
        SP-gra2seq & $0.15\pm0.49$ & $0.19\pm0.56$\\
        DC-gra2seq & $0.22\pm0.68$ & $0.23\pm0.61$\\
        SketchEdit & $1.20\pm0.86$ & $1.47\pm0.82$\\
        SketchEdit$^+$ & $1.04\pm0.85$ & $0.32\pm0.67$\\
        Sketch-HARP & $0.25\pm0.67$ & $0.37\pm0.75$\\
        SketchMod & $\pmb{1.74}\pm\pmb{0.86}$ & $\pmb{1.68}\pm\pmb{0.86}$\\
        \bottomrule
    \end{tabular}
    \caption{Human scoring results (mean$\pm$std.) on stroke expansion and stroke replacement.}
    \label{tab:human_scoring}
\end{table}

Besides, we also present some qualitative results on both stroke expansion and replacement in Fig.~\ref{fig:edit_results}. We can figure out that the sketches edited by SketchMod are the best in both recognition and semantic representation. For example, in the 1-st column in stroke replacement, SketchMod resizes the bee's body to align with the lengths of stripes, and in 3-rd column in stroke expansion, we reposition the same right eye from the pig face to become the left eye. Besides, we also yield some interesting results, e.g., in the 2-nd column in stroke expansion, the wing from the bee is transplanted on the giraffe's back, contributing to a ``Pegasus''.

\subsection{Manipulating Sketch Synthesis at Stroke-Level via Editable Attributes}

\begin{figure}[!t]
    \centering
    \includegraphics[width=0.98\columnwidth]{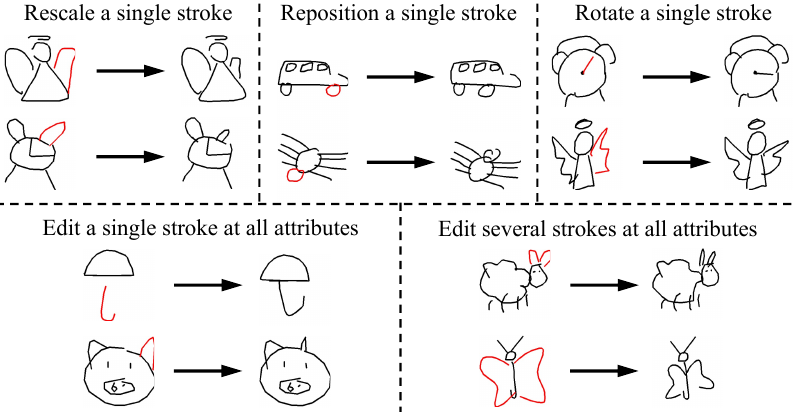}
    \caption{Manipulating sketch synthesis at stroke-level by SketchMod.}
    \label{fig:adjust_attributes}
\end{figure}

In SketchMod's sketch generator, the attributes of strokes are stored in the exposed vectors $\bm p$ in Eq.~(\ref{eq:mixing_stroke}), which makes it convenient to manipulate the scale, orientation and position of a stroke directly by adjusting their values. Fig.~\ref{fig:adjust_attributes} shows some exemplary results.

As shown in Fig.~\ref{fig:adjust_attributes}, SketchMod supports both single- and multi-stroke editing through parametric transformations at scale, orientation and position, enabling either individual or simultaneous manipulations of stroke attributes. Accordingly, with a given sketch, it is flexible to modify some selected strokes to achieve desired results.

\subsection{Sketch Reconstruction}

We further evaluate each model' sketch generative performance on sketch reconstruction. No noise is applied on sketches, and models are required to generate the reconstructed sketches with as much details as possible from the input. Table~\ref{tab:reconstruction} reports the quantitative results on sketch reconstruction. Qualitative comparisons are shown in Appendix D.

According to Table~\ref{tab:reconstruction}, our SketchMod achieves the best sketch reconstruction performance among all models. By enabling the message communication among strokes via the self-attention network $\xi$ in Eq.~(\ref{eq:mixing_stroke}), a stroke could receives the features passing from its neighboring strokes, yielding comprehensive stroke representation to benefit sketch reconstruction. 

\subsection{Impact of Learning Offset Attributes}\label{sec:ablation}

In this section, we discuss whether the offset attributes between strokes are beneficial in source stroke refinement. Table~\ref{tab:ablation1} reports the sketch edit results by three variants of SketchMod. ``Attribute $\times$, Offset $\times$'' denotes a variant that refines the source stroke directly from the learned stroke embeddings, i.e., replacing Eq.~(\ref{eq:predictor}) by $\hat{\bm e}_\sigma=\psi(\bm E)$, where $\psi$ reduces to a vanilla self-attention network. ``Attribute $\checkmark$, Offset $\times$'' realizes the source stroke refinement by learning from stroke attributes instead of their offsets, i.e., reforming Eq.~(\ref{eq:predictor}) as $\hat{\bm e}_\sigma=\psi(\bar{\bm E}+\bm P)$, where $\bm P=\text{CAT}(\bm p_\sigma; \{\bm p_k\})$ and $\psi$ is a vanilla self-attention network as well. ``Attribute $\times$, Offset $\checkmark$'' denotes the proposed model.

All three models aggregate messages from strokes in the target sketch to enhance the source stroke via attention-based networks for refinement. Compared with SketchMod which directly encodes stroke relationships by their offset attributes to align the source stroke with the target's patterns, both variants never guide networks to quantify the hidden relative positions between strokes. As a result, the proposed SketchMod achieves the best sketch edit performance.

\begin{table}[!t]
    \centering
    \small
    \begin{tabular}{@{}clcrcc@{}}
        \toprule
        Dataset & Model & $Rec$ & FID & LPIPS & CLIP-S\\
        \midrule
		\multirow{10}*{DS1} & RPCL-pix2seq & 81.80 & 22.74 & 0.37 & 88.57\\
		~ & RPCL-pix2seqH & 87.82 & 17.30 & 0.38 & 88.66\\
		~ & SketchHealer & 87.03 & 27.72 & 0.39 & 88.17 \\
		~ & Lmser-pix2seq & 90.50 & 18.94 & 0.36 & 89.74 \\
		~ & SP-gra2seq & 89.83 & 10.80 & 0.38 & 88.72 \\
		~ & DC-gra2seq & 90.41 & 12.83 & 0.30 & 93.84\\
        ~ & SketchEdit & 87.09 & 3.11 & 0.11 & 96.73 \\
        ~ & SketchEdit$^+$ & 86.16 & 3.71 & 0.12 & 96.35\\
        ~ & Sketch-HARP & 89.90 & 9.96 & 0.28 & 91.60 \\
        ~ & SketchMod & \pmb{90.97} & \pmb{2.48} & \pmb{0.10} & \pmb{97.30} \\
        \midrule
        \multirow{10}*{DS2} & RPCL-pix2seq & 74.94 & 36.56 & 0.28 & 85.47 \\
		~ & RPCL-pix2seqH & 84.40 & 26.90 & 0.34 & 88.13 \\
		~ & SketchHealer & 74.93 & 31.02 & 0.29 & 91.46 \\
		~ & Lmser-pix2seq & 85.02 & 10.10 & 0.32 & 91.10 \\
		~ & SP-gra2seq & 76.89 & 34.06 & 0.45 & 85.17 \\
		~ & DC-gra2seq & 85.67 & 11.01 & 0.30 & 94.46 \\
        ~ & SketchEdit & 87.16 & 5.50 & 0.17 & 94.68\\
        ~ & SketchEdit$^+$ & 86.62 & 6.66 & 0.18 & 94.49\\
        ~ & Sketch-HARP & 87.71 & 6.97 & 0.22 & 93.31 \\
        ~ & SketchMod & \pmb{90.66} & \pmb{3.66} & \pmb{0.13} & \pmb{96.12} \\
        \bottomrule
    \end{tabular}
    \caption{Comparisons on sketch reconstruction.}
    \label{tab:reconstruction}
\end{table}

\begin{table}[!t]
    \centering
    \small
    \begin{tabular}{@{}ccccccc@{}}
        \toprule
        Dataset & Attribute & Offset & $Rec$ & FID & LPIPS & CLIP-S\\
        \midrule
		\multirow{3}*{DS1} & $\times$ & $\times$ & 71.28 & 7.37 & 0.25 & 93.65\\
        ~ & $\checkmark$ & $\times$ & 56.52 & 8.22 & 0.23 & 93.32\\
        ~ & $\times$ & $\checkmark$ & \pmb{71.51} & \pmb{3.44} & \pmb{0.19} & \pmb{95.52} \\
        \midrule
        \multirow{3}*{DS2} & $\times$ & $\times$ & 87.70 & 8.17 & 0.24 & 93.41\\
        ~ & $\checkmark$ & $\times$ & 77.55 & 7.93 & 0.23 & 93.55 \\
        ~ & $\times$ & $\checkmark$ & \pmb{89.25} & \pmb{3.86}  & \pmb{0.17}  & \pmb{95.43} \\
        \bottomrule
    \end{tabular}
    \caption{Comparisons between variants of SketchMod on sketch edit, by refining the source stroke through different approaches.}
    \label{tab:ablation1}
\end{table}

\section{Conclusions}

We have proposed SketchMod for flexible stroke-level sketch edit by refining the source stroke through transformation to align it with the target sketch's patterns. Three key stroke attributes are captured and utilized to learn the relative displacements from the source stroke to another in the target sketch. By equipping stroke embeddings with their relative positions though offset attributes, we aggregate messages passing from the target sketch to obtain the refined source stroke, further realizing sketch edit at stroke-level in a flexible manner.

\begin{figure*}[t!]
    \centering
    \includegraphics[width=1.95\columnwidth]{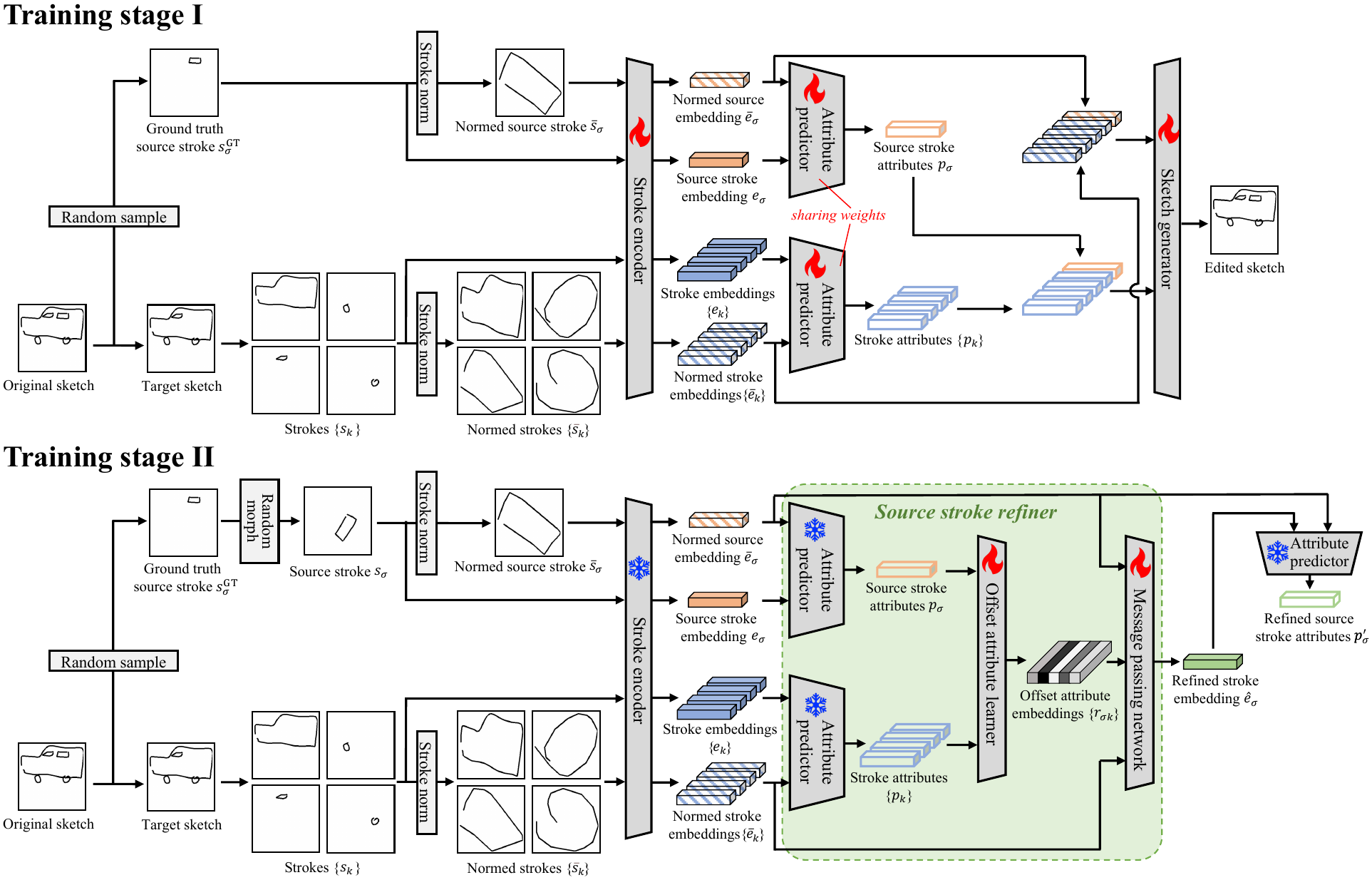}
    \caption{An overview about the two training stages of SketchMod.}
    \label{fig:training_stages}
\end{figure*}

\section*{Appendix}

\appendix

\section{Details about Training a SketchMod}

\subsection{The Pipeline of Network Training}

Fig.~\ref{fig:training_stages} shows an overview about the two training stages of the proposed SketchMod.

In stage I, we aim to train the stroke encoder, the attribute predictor and the sketch generator via sketch reconstruction. For a sketch input, we randomly select a stroke and regard it as the ground truth source stroke $\bm s_\sigma^\text{GT}$. The rest strokes $\{\bm s_k\}$ in the original sketch are as the target sketch. Both $\bm s^\text{GT}_\sigma$ and $\{\bm s_k\}$ are normalized to yield their corresponding normalized strokes $\bar{\bm s}_\sigma$ and $\{\bar{\bm s}_k\}$, respectively, which are fed into a stroke encoder to capture their embeddings $\bar{\bm e}_\sigma$ and $\{\bar{\bm e}_k\}$. Each pair of $\bm e$ and $\bar{\bm e}$ is sent into an attribute predictor to predict its stroke attributes $\bm p$. Finally, we collect all normalized stroke embeddings with their predicted stroke attributes, and utilize them for sketch reconstruction via a sketch generator. The objective in this stage is in Eq.~(\ref{eq:stage1_loss}) in the manuscript.

And in stage II, leveraging the pretrained parameters, we keep the stroke encoder, the attribute predictor and the sketch generator frozen, and target to train the offset attribute learner and the message passing network in the source stroke refiner. We apply random noises over the attributes of $\bm s_\sigma^\text{GT}$ to obtain the input source stroke $\bm s_\sigma$ for sketch edit. Both the predicted attributes captured from $\bm p_\sigma$ and $\{\bm p_k\}$ from the target sketch are utilized to learn offset attribute embeddings $\{\bm r_{\sigma k}\}$ from stroke $\bm s_k$ to $\bm s_\sigma$. A message passing network collects all the normalized stroke embeddings and aggregates the messages passing from $\bm s_k$ to enhance $\bm s_\sigma$, to finally obtain the refined source stroke embedding $\hat{\bm e}_\sigma$. The pretrained attribute predictor then computes its refined stroke attributes as $\bm p_\sigma^{\prime}$. The objective in this stage is in Eq.~(\ref{eq:stage2_loss}) in the manuscript.

When training a SketchMod, the AdamW optimizer parameterized by $\beta_1=0.9$, $\beta_2=0.999$ and $\epsilon=10^{-8}$ is employed. Following \cite{li2024sketchedit}, the CosineAnnealingLR scheduler \cite{smith2019super} with the peak learning rates as 0.001 is utilized in both training stages of SketchMod. We train our model on a single NVIDIA RTX 4090 D, and set the mini-batch size at 80.

\subsection{How to Get a Corrupted Source Stroke in SketchMod Training}

During SketchMod training, we corrupt the ground truth source stroke by applying noises on its stroke attributes. Then the noised source stroke is regarded as the source stroke input in SketchMod training. In this way, we can achieve the ground truth results for both source stroke refinement and sketch edit to guide the network training.

\begin{figure*}[!t]
    \centering
    \includegraphics[width=1.8\columnwidth]{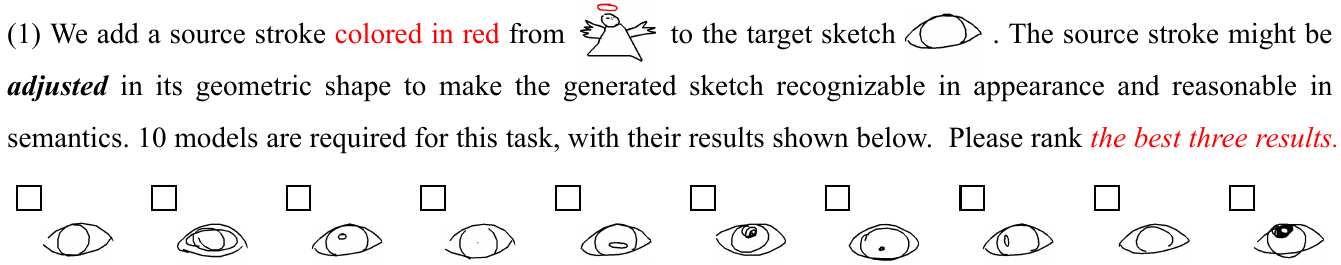}
    \caption{An exemplary question for stroke expansion in the questionnaire.}
    \label{fig:questionnaire_expansion}
\end{figure*}

\begin{figure*}[!t]
    \centering
    \includegraphics[width=1.8\columnwidth]{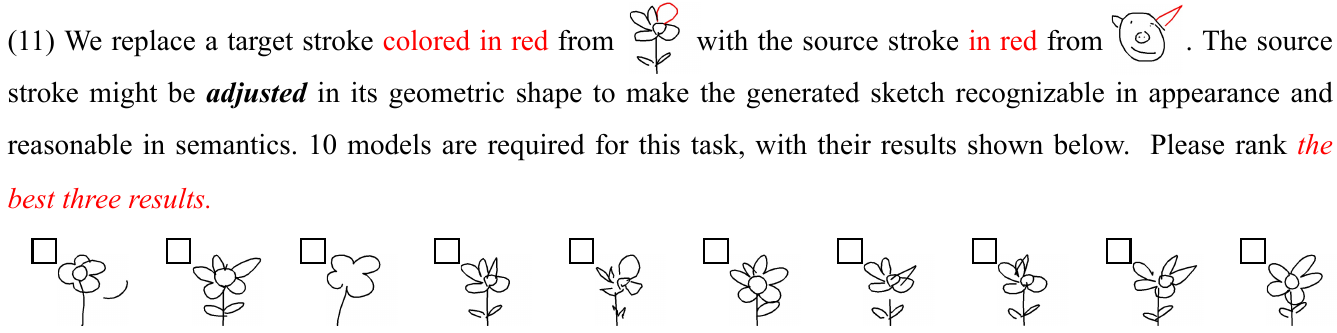}
    \caption{An exemplary question for stroke replacement in the questionnaire.}
    \label{fig:questionnaire_replacement}
\end{figure*}

When a ground truth source stroke $\bm s_\sigma^\text{GT}$ is picked, we firstly normalize it to obtain its corresponding normalized stroke $\bar{\bm s}_\sigma$ and the ground truth source stroke attributes $\bm p_\sigma^\text{GT}=[a^\text{GT}_\sigma,b^\text{GT}_\sigma,\theta^\text{GT}_\sigma,\log(\bm\tau^\text{GT}_\sigma)]$, where $(a^\text{GT}_\sigma,b^\text{GT}_\sigma)$, $\theta^\text{GT}_\sigma$ and $\bm\tau^\text{GT}_\sigma$ denote its starting position on the canvas, its orientation by a rotation angle and its two-dimensional scales. We corrupt them by,
\begin{align}
    &(a^\text{N}_\sigma,b^\text{N}_\sigma) = (a^\text{GT}_\sigma+\epsilon_a,b^\text{GT}_\sigma+\epsilon_b),\>\epsilon_a,\epsilon_b\sim\mathcal{U}(\epsilon|-1, 1),\\
    &\theta^\text{N}_\sigma = \theta^\text{GT}_\sigma+\epsilon_\theta,\quad\quad\quad\quad\epsilon_\theta\sim\mathcal{U}(\epsilon|-\frac{\pi}{2},\frac{\pi}{2}),\\
    &\bm\tau^\text{N}_\sigma = \bm\tau^\text{GT}_\sigma \odot[\epsilon_{\tau 1}, \epsilon_{\tau 2}]^\top, \>\epsilon_{\tau 1},\epsilon_{\tau 2}\sim\mathcal{U}(\epsilon|0.3,2.2),
\end{align}
where $\odot$ denotes the element-wise production. With the corrupted attributes $\bm p^\text{N}_\sigma=[a^\text{N}_\sigma,b^\text{N}_\sigma,\theta^\text{N}_\sigma,\log(\bm\tau^\text{N}_\sigma)]$, we apply it to the normalized stroke $\bar{\bm s}_\sigma$ through transformation to obtain the input source stroke $\bm s_\sigma$ in SketchMod training.

\section{How to Evaluate Human Perception Quality for Stroke-Level Sketch Edit}

\subsection{Questionnaire Details}

The questionnaire consists of 20 questions, with 10 sketch expansion and 10 sketch replacement questions, respectively. Fig.~\ref{fig:questionnaire_expansion} and Fig.~\ref{fig:questionnaire_replacement} show two exemplary questions for each type, respectively. 

In the questions for stroke expansion, a stroke highlighted in red, regarded as the source stroke, is required to be expanded onto the target sketch. And in the questions for stroke replacement, for a given target sketch, the selected stroke in red would be replaced by the source stroke collected from another sketch. Each edited sketch is fed into 10 models respectively as a reference, and models are required to generate sketches to preserve detailed patterns from the given input. In a question, human volunteers are asked to select and rank the best three generated sketches.

\subsection{Scoring Rules and Human Perception Evaluation}

For the 10 generated sketches (by SketchMod and 9 baseline models) in each question, the best three are selected and ranked by each human subject. More specifically, for each question, a human subject values the best three results with scores 3, 2 and 1, respectively, leaving the rest results with 0 scores. Totally 42 volunteers have participated in this evaluation, and we have 10 questions per edit type (stroke expansion and replacement). $42\times 10=420$ scores are collected for each question type and we compute the mean values and standard deviations, respectively, which are reported in Table 1 in the manuscript.

\section{Why Predicts Stroke Attributes, Since the Ground Truth Attributes Could be Obtained via Stroke Normalization?}

When training our SketchMod, in the stage I in Fig.~\ref{fig:training_stages}(top), we introduce two attribute predictors with shared weights to predict the stroke attributes $\bm p_\sigma$ and $\{\bm p_k\}$ from the source stroke and the ones in the target sketch, respectively. These predicted attributes further cooperate with their corresponding normalized stroke embeddings $\bar{\bm e}_\sigma$ and $\{\bar{\bm e}_k\}$ for sketch generation. Since the ground truth stroke attributes $\bm p^\text{GT}_\sigma$ and $\{\bm p_k^\text{GT}\}$ have already been computed during the stroke normalization via Eq.~(\ref{eq:normalization}) in the manuscript, is there any reason which prevents us to directly utilize these $\bm p^\text{GT}_\sigma$ and $\{\bm p_k^\text{GT}\}$ for sketch generation? Here follows the reason.

In SketchMod's stage II in Fig.~\ref{fig:training_stages}(bottom) for stroke-level sketch edit, we aim to refine the source stroke and capture its refined stroke attributes $\bm p^{\prime}_\sigma$ to feed it into the sketch generator. By learning the offset attribute embeddings between strokes, we are able to yield a refined source stroke embedding $\hat{\bm e}_\sigma$, instead of a refined source stroke realized by a sequence of drawing actions. Accordingly, we CANNOT compute its new refined stroke attributes by stroke normalization via Eq.~(\ref{eq:normalization}) in the manuscript. One practical idea is to build an attribute predictor to learn the attributes $\bm p^{\prime}_\sigma$ by feeding the predictor with the refined stroke embedding $\hat{\bm e}_\sigma$, just as what we made in SketchMod. But if we employ the attribute predictor only in training stage II, we might not have enough paired data (the predicted $\bm p^{\prime}_\sigma$ and its corresponding ground truth $\bm p^\text{GT}_\sigma$) to obtain a well-trained predictor. It is because for each sketch sample, we only have one source stroke with a single pair of $\bm p^{\prime}_\sigma$ and $\bm p^\text{GT}_\sigma$. Hence, it is more practical to train the attribute predictor in training stage I, since we would have $K+1$ pairs of data $\bm p_{\{\sigma, k\}}$ and $\bm p^\text{GT}_{\{\sigma, k\}}$ for each sketch. With a well-trained attribute predictor through the training stage I, SketchMod could predict an accurate $\bm p^{\prime}_\sigma$ in training stage II with the refined embedding $\hat{\bm e}_\sigma$, arriving at efficient learning of the source stroke refinement.

Besides, in Sect.~\ref{sec:ablation} in the manuscript, we introduce an ablation study in Table~\ref{tab:ablation2} to draw the conclusion that feeding the sketch generator with normalized stroke embeddings $\bar{\bm e}_{\{\sigma, k\}}$ with their corresponding predicted attributes $\bm p_{\{\sigma, k\}}$ (corresponding to the proposed SketchMod) could achieve better sketch edit performance than feeding the generator with the (predicted) stroke embeddings $\hat{\bm e}_\sigma$ and $\{\bm e_k\}$ (corresponding to the variant ``Attribute $\times$, Offset $\times$''). Thus, we prefer to 
capture the refined stroke's attributes, instead of sending it into the sketch generator.

\section{Qualitative Comparisons on Sketch Reconstruction}

\begin{figure}[!h]
    \centering
    \includegraphics[width=\columnwidth]{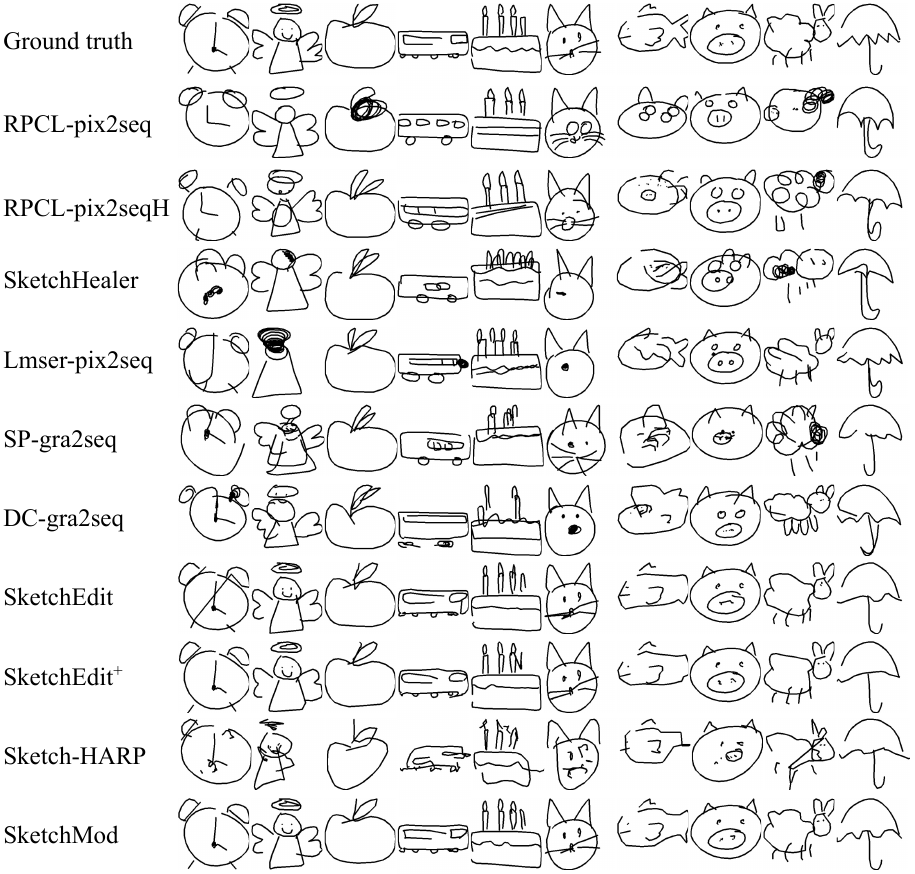}
    \caption{Exemplary results on sketch reconstruction.}
    \label{fig:reconstruction}
\end{figure}

Fig.~\ref{fig:reconstruction} shows the qualitative results generated by our SketchMod and other nine baselines on sketch reconstruction. Our SketchMod preserves is powerful at preserving sketch details from the input. For example, in the 8-th column, we generate two clear nostrils of the pig, while SketchEdit fails. The other baselines, e.g., RPCL-pix2seq and Lmser-pix2seq, which also generate two clear nostrils, preserve less details from the ground truth sketch.

\section{Ablation Study: Training a SketchMod in Multiple Stages or Not}

In this section, we further discuss the advantages by learning SketchMod via two separated stages. A variant of SketchMod is introduced, which trains all the network parameters through a single stage with the objective as,
\begin{align}
    \min\limits_{f, \mathcal{F}, h, \psi, \xi, \mathcal{G}_{\text{seq}}, \mathcal{G}_{\text{img}}}\mathcal{L}=&\mathcal{L}_{\text{seq}}+0.2\cdot\|\bm I-\hat{\bm I}\|_2^2+\sum\nolimits_k\|\bm p^{\text{GT}}_k-\bm p_k\|_2^2\nonumber\\
    &+\|\text{sg}(\bm e^{\text{GT}}_\sigma)-\hat{\bm e}_\sigma\|_2^2+\|\bm p^{\text{GT}}_\sigma-\bm p^{\prime}_\sigma\|_2^2.
\end{align}

Table~\ref{tab:ablation2} reports the comparisons between two types of SketchMod on sketch edit. And the experimental results reveal that training SketchMod in two separated stages would achieve better sketch edit performance.

\begin{table}[!t]
    \centering
    \small
    \begin{tabular}{@{}ccrrrr@{}}
        \toprule
        Dataset & \makecell{Multiple\\training stages} & $Rec$ & FID & LPIPS & CLIP-S\\
        \midrule
		\multirow{2}*{DS1} & $\times$ & 63.37 & 17.47 & 0.29 & 92.22 \\
        ~ & $\checkmark$ & \pmb{71.51} & \pmb{3.44} & \pmb{0.19} & \pmb{95.52} \\
        \midrule
        \multirow{2}*{DS2} & $\times$ & 83.71 & 19.26 & 0.28 & 91.84\\
        ~ & $\checkmark$ & \pmb{89.25} & \pmb{3.86} & \pmb{0.17}  & \pmb{95.43} \\
        \bottomrule
    \end{tabular}
    \caption{Comparisons between two variants of SketchMod on sketch edit, by training the network via a single stage or via two stages.}
    \label{tab:ablation2}
\end{table}

\section*{Acknowledgments}

This work was supported by the National Natural Science Foundation of China (62406064, 62477006), and the Fundamental Research Funds for the Central Universities (2232024D-28).

\bibliographystyle{named}
\bibliography{ijcai26}

\end{document}